\documentclass[10pt,twocolumn,letterpaper]{article}

\usepackage{cvpr}              

\usepackage{graphicx}
\usepackage{amsmath}
\usepackage{amssymb}
\usepackage{booktabs}
\usepackage{tabularx}
\usepackage{multirow}
\usepackage{makecell}

\usepackage[pagebackref,breaklinks,colorlinks]{hyperref}

\usepackage[capitalize]{cleveref}
\crefname{section}{Sec.}{Secs.}
\Crefname{section}{Section}{Sections}
\Crefname{table}{Table}{Tables}
\crefname{table}{Tab.}{Tabs.}

\def\cvprPaperID{11345} 
\def\confName{CVPR}
\def\confYear{2025}
\begin{document}

\title{Hallo3: Highly Dynamic and Realistic Portrait Image Animation with Video Diffusion Transformer}
\author{Jiahao Cui\textsuperscript{1},
        Hui Li\textsuperscript{1},
        Yun Zhan\textsuperscript{1},
        Hanlin Shang\textsuperscript{1},
        Kaihui Cheng\textsuperscript{1},
        Yuqi Ma\textsuperscript{1},
        Shan Mu\textsuperscript{1},
        Hang Zhou\textsuperscript{2},\\
        Jingdong Wang\textsuperscript{2},\;\;\;
        Siyu Zhu\textsuperscript{1,3} \\
\textsuperscript{1}Fudan University, \;\;\textsuperscript{2}Baidu Inc, \;\;\textsuperscript{3}Shanghai Academy of AI for Science\\
\url{https://fudan-generative-vision.github.io/hallo3}
}
\maketitle

\begin{abstract}
Existing methodologies for animating portrait images face significant challenges, particularly in handling non-frontal perspectives, rendering dynamic objects around the portrait, and generating immersive, realistic backgrounds.
In this paper, we introduce the first application of a pretrained transformer-based video generative model that demonstrates strong generalization capabilities and generates highly dynamic, realistic videos for portrait animation, effectively addressing these challenges.
The adoption of a new video backbone model makes previous U-Net-based methods for identity maintenance, audio conditioning, and video extrapolation inapplicable. To address this limitation, we design an identity reference network consisting of a causal 3D VAE combined with a stacked series of transformer layers, ensuring consistent facial identity across video sequences.
Additionally, we investigate various speech audio conditioning and motion frame mechanisms to enable the generation of continuous video driven by speech audio. Our method is validated through experiments on benchmark and newly proposed wild datasets, demonstrating substantial improvements over prior methods in generating realistic portraits characterized by diverse orientations within dynamic and immersive scenes.
\end{abstract}

\begin{figure*}
    \centering
    \includegraphics[width=1\linewidth]{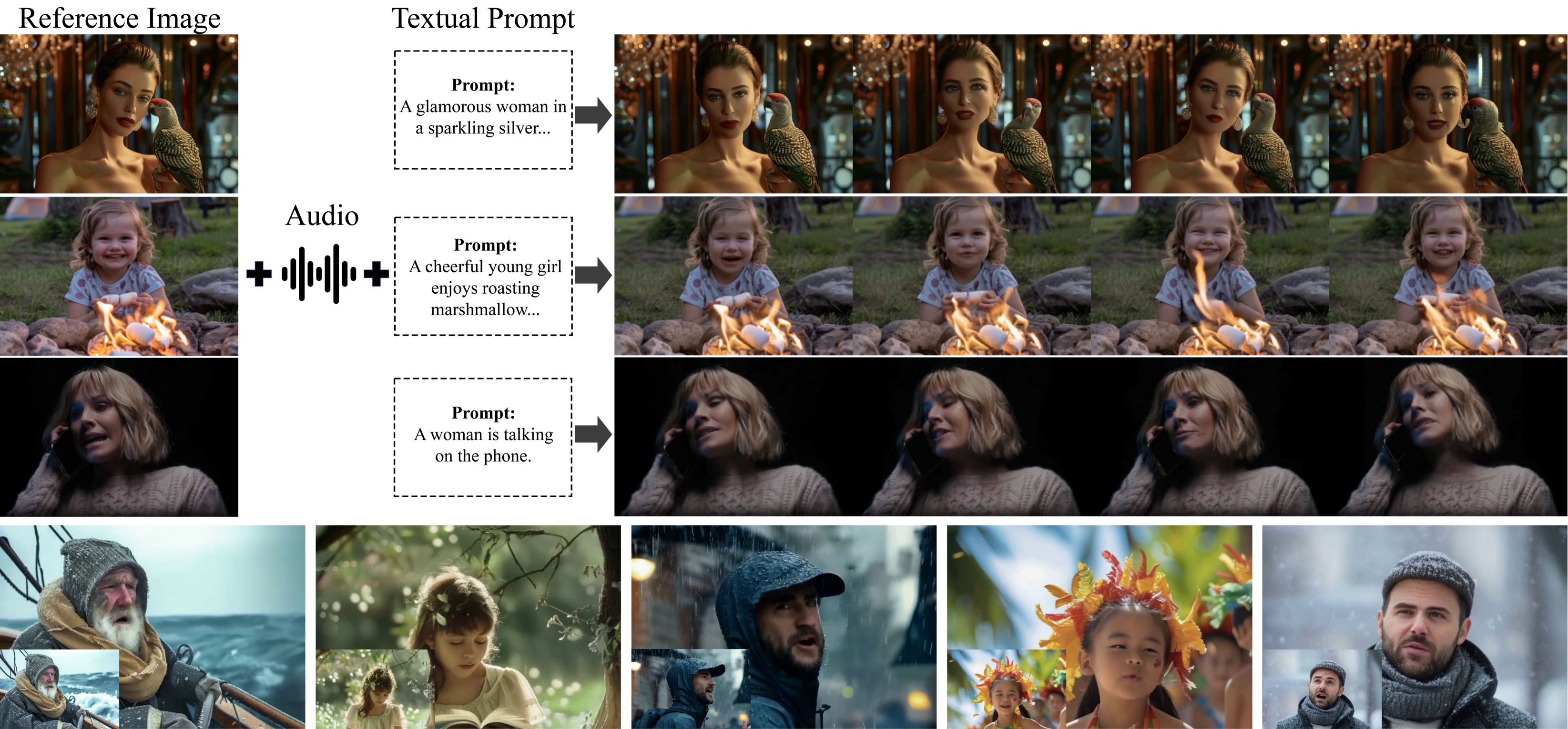}
    \caption{Demonstration of the proposed approach. Given a reference image, an audio sequence, and a textual prompt, the method generates animated portraits from frontal or different perspectives while preserving the portrait identity over extended durations. Additionally, it incorporates dynamic foreground and background elements, with temporal consistency and high visual fidelity. }
    \label{fig:enter-label}
    \vspace{-5mm}
\end{figure*}

\section{Introduction}
Portrait image animation refers to the process of generating realistic facial expressions, lip movements, and head poses based on portrait images. 
This technique leverages various motion signals, including audio, textual prompts, facial keypoints, and dense motion flow. 
As a cross-disciplinary research task within the realms of computer vision and computer graphics, this area has garnered increasing attention from both academic and industrial communities. 
Furthermore, portrait image animation has critical applications across several sectors, including film and animation production, game development, social media content creation, and online education and training.

In recent years, the field of portrait image animation has witnessed rapid advancements. 
Early methodologies predominantly employed facial landmarks—key points~\cite{siarohin2019first,zakharov2020fast,zhang2022sadtalker} on the face utilized for the localization and representation of critical regions such as the mouth, eyes, eyebrows, nose, and jawline. 
Additionally, these methods~\cite{gao2023high,ren2021pirenderer,zhang2023metaportrait} incorporated 3D parametric models, notably the 3D Morphable Model (3DMM)~\cite{blanz2003face}, which captures variability in human faces through a statistical shape model integrated with a texture model. 
However, the application of explicit approaches grounded in intermediate facial representations is constrained by the accuracy of expression and head pose reconstruction, as well as the richness and precision of the resultant expressions.
Simultaneously, significant advancements in Generative Adversarial Networks (GANs) and diffusion models have notably benefited portrait image animation. 
These advancements~\cite{corona2024vlogger,liu2024anitalker,tian2024emo,wei2024aniportrait,xu2024vasa} enhance the high-resolution and high-quality generation of realistic facial details, facilitate generalized character animation, and enable long-term identity preservation. 
Recent contributions to the field—including Live Portrait~\cite{guo2024liveportrait}, which leverages GAN technology for portrait animation with stitching and retargeting control, as well as various end-to-end methods such as VASA-1~\cite{xu2024vasa}, EMO~\cite{tian2024emo}, and Hallo~\cite{xu2024hallo,cui2024hallo2} employing diffusion models—exemplify these advancements.

Despite these improvements, existing methodologies encounter substantial limitations. 
First, many current facial animation techniques emphasize eye gaze, lip synchronization, and head posture while often depending on reference portrait images that present a frontal, centered view of the subject. 
This reliance presents challenges in handling profile, overhead, or low-angle perspectives for portrait animation. Secondly, accounting for significant accessories, such as holding a smartphone, microphone, or wearing closely fitted objects, presents challenges in generating realistic motion for the associated objects within video sequences.
Third, existing methods often assume static backgrounds, undermining their ability to generate authentic video effects in dynamic scenarios, such as those with campfires in the foreground or crowded street scenes in the background.

Recent advancements in diffusion transformer-based video generation models~\cite{yang2024cogvideox, polyak2024movie, bao2024vidu, liu2024sora} have addressed several challenges associated with traditional video generation techniques, including issues of realism, dynamic movement, and subject generalization. 
In this paper, we present the first application of a pretrained DiT-based video generative model to the task of portrait image animation.
The introduction of this new video backbone model renders previous U-Net-based methods for identity maintenance, audio conditioning, and video extrapolation impractical. 
We tackle these issues from three distinct perspectives.
(1)~\textbf{Identity preservation}: We employ a 3D VAE in conjunction with a stack of transformer layers as an identity reference network, enabling the embedding and injection of identity information into the denoising latent codes for self-attention. This facilitates accurate representation and long-term preservation of the facial subject's identity.
(2)~\textbf{Speech audio conditioning}: We achieve high alignment between speech audio—serving as motion control information—and facial expression dynamics during training, which allows for precise control during inference. We investigate the use of adaptive layer normalization and cross-attention strategies, effectively integrating audio embeddings through the latter.
(3)~\textbf{Video extrapolation}: Addressing the limitations of the DiT model in generating contineous videos, which is constrained to a maximum of several tens of frames, we propose a strategy for long-duration video extrapolation. 
This approach uses motion frames as conditional information, wherein the final frames of each generated video serve as inputs for subsequent clip generation.

We validate our approach using benchmark datasets, including HTDF and Celeb-V, demonstrating results comparable to previous methods that are constrained to limited datasets characterized by frontal, centered faces, static backgrounds, and defined expressions. 
Furthermore, our method successfully generates dynamic foregrounds and backgrounds, accommodating complex poses, such as profile views or interactions involving devices like smartphones and microphones, yielding realistic and smoothly animated motion, thereby addressing challenges that previous methodologies have struggled to resolve effectively. 
\section{Related Work}
\noindent\textbf{Portrait Image Animation.}
Recent advancements in the domain of portrait image animation have been significantly propelled by innovations in audio-driven techniques. 
Notable frameworks, such as LipSyncExpert~\cite{prajwal2020lip} and SadTalker~\cite{zhang2023sadtalker}, have tackled challenges related to facial synchronization and expression modulation, achieving dynamic lip movements and coherent head motions. 
Concurrently, DiffTalk~\cite{shen2023difftalk} and VividTalk~\cite{sun2023vividtalk} have integrated latent diffusion models, enhancing output quality while generalizing across diverse identities without the necessity for extensive fine-tuning. 
Furthermore, studies such as DreamTalk~\cite{ma2023dreamtalk} and EMO~\cite{tian2024emo} underscore the importance of emotional expressiveness by showcasing the integration of audio cues with facial dynamics. 
AniPortrait~\cite{wei2024aniportrait} and VASA~\cite{xu2024vasa} propose methodologies that facilitate the generation of high-fidelity animations, emphasizing temporal consistency along with effective exploitation of static images and audio clips. 
In addition, recent innovations like LivePortrait~\cite{guo2024liveportrait} and Loopy~\cite{jiang2024loopy} focus on enhancing computational efficiency while ensuring realism and fluid motion. 
Furthermore, the works of Hallo~\cite{xu2024hallo} and Hallo2~\cite{cui2024hallo2} have made significant progress in extending capabilities to facilitate long-duration video synthesis and integrating adjustable semantic inputs, thereby marking a step towards richer and more controllable content generation. 
Nevertheless, existing facial animation techniques still encounter limitations in addressing extreme facial poses, accommodating background motion in dynamic environments, and incorporating camera movements dictated by textual prompts.

\noindent\textbf{Diffusion-Based Video Generation.}
Unet-based diffusion video generation has made notable strides, exemplified by frameworks such as Make-A-Video and MagicVideo~\cite{zhou2022magicvideo}. 
Specifically, Make-A-Video~\cite{singer2022make} capitalizes on pre-existing T2I models to enhance training efficiency without necessitating paired text-video data, thereby achieving state-of-the-art results across a variety of qualitative and quantitative metrics. 
Simultaneously, MagicVideo~\cite{zhou2022magicvideo} employs an innovative 3D U-Net architecture to operate within a low-dimensional latent space, achieving efficient video synthesis while significantly reducing computational requirements. 
Building upon these foundational principles, AnimateDiff~\cite{guo2023animatediff} introduces a motion module that integrates seamlessly with personalized T2I models, allowing for the generation of temporally coherent animations without the need for model-specific adjustments. 
Additionally, VideoComposer~\cite{wang2023videocomposer} enhances the controllability of video synthesis by incorporating spatial, temporal, and textual conditions, which facilitates improved inter-frame consistency. 
The development of diffusion models continues with the advent of DiT-based approaches such as CogVideoX~\cite{yang2024cogvideox} and Movie Gen~\cite{polyak2024movie}. 
CogVideoX employs a 3D Variational Autoencoder to improve video fidelity and narrative coherence, whereas Movie Gen establishes a robust foundation for high-quality video generation complemented by advanced editing capabilities. 
In the present study, we adopt the DiT diffusion formulation to optimize the generalization capabilities of the generated video.


\begin{figure*}[t!]
    \centering

    \begin{minipage}[b]{0.65\textwidth} 
        \centering
        \includegraphics[width=\linewidth]{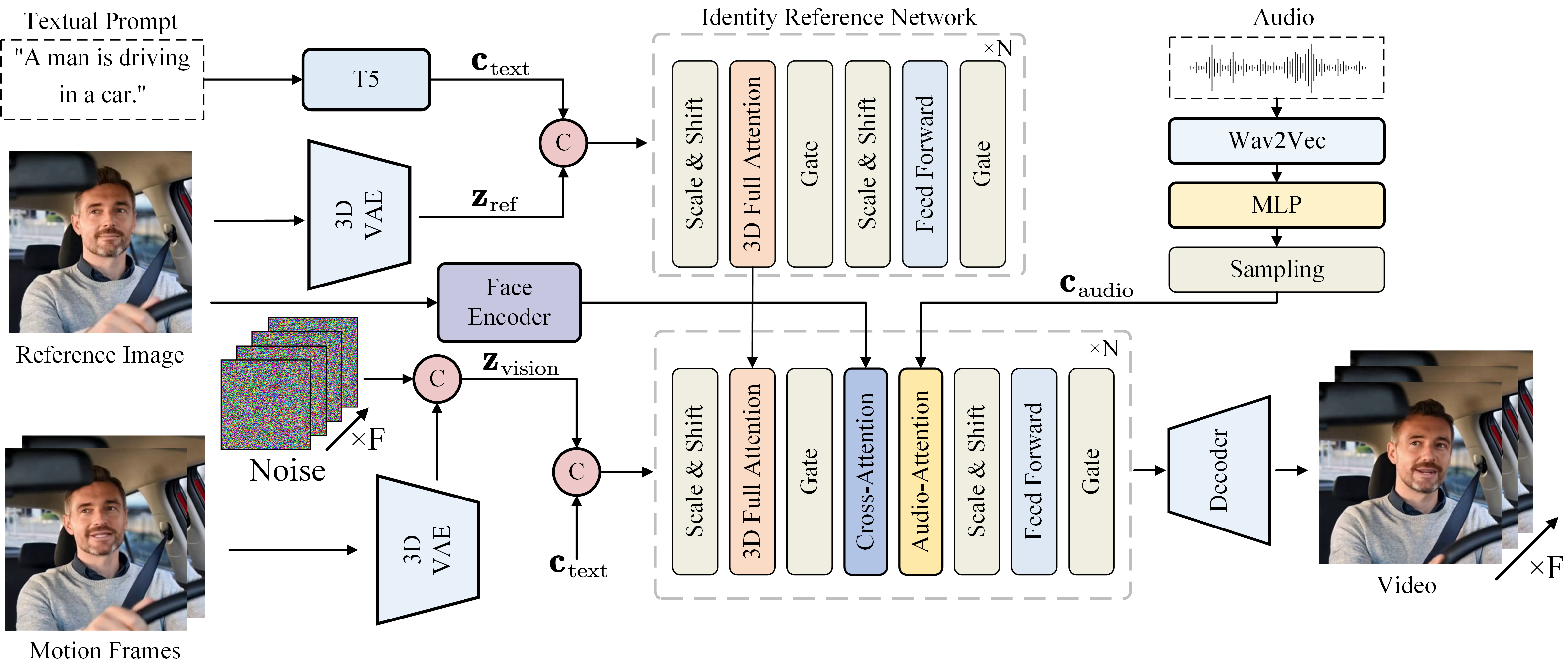} 
        \caption{The overview of the proposed method. Specifically, the method takes a reference image, an audio sequence, and a textual prompt as inputs to generate a video output with temporal consistency and high visual fidelity. We leverage the casual 3D VAE, T5, and Wav2Vec models to process the visual, textual, and audio features, respectively. The Identity Reference Network extracts identity features from the input reference image and textual prompt, enabling controllable animation while preserving the subject's appearance. The audio encoder generates motion information for lip synchronization, while the face encoder extracts facial features to maintain consistency in facial expressions. The 3D Full Attention and Audio-Attention Modules combine identity and motion data within a denoising network, producing high-fidelity, temporally consistent, and controllable animated videos.}
        \label{fig:architecture}

    \end{minipage}
    \hfill
    \begin{minipage}[b]{0.32\textwidth} 
        \centering
        \begin{minipage}[b]{\textwidth} 
                \centering
                \begin{subfigure}{0.3\linewidth} 
                    \centering
                    \includegraphics[width=\linewidth]{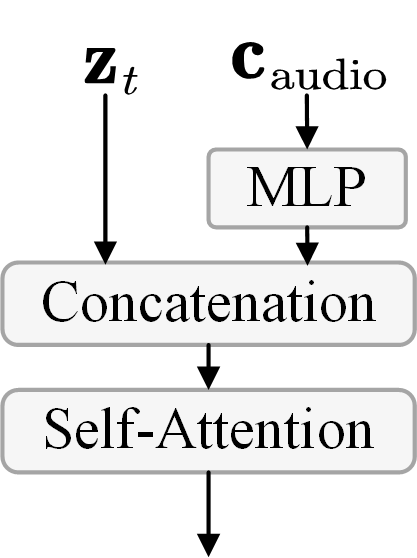} 
                    \caption{}  
                    \label{fig:AudioConditionSA}  
                \end{subfigure}%
                \hfill
                \begin{subfigure}{0.3\linewidth} 
                    \centering
                    \includegraphics[width=\linewidth]{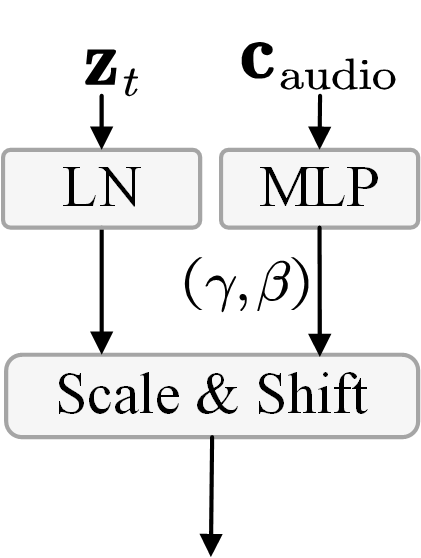}  
                    \caption{}  
                    \label{fig:AudioConditionAda}  
                \end{subfigure}%
                \hfill
                \begin{subfigure}{0.3\linewidth} 
                    \centering
                    \includegraphics[width=\linewidth]{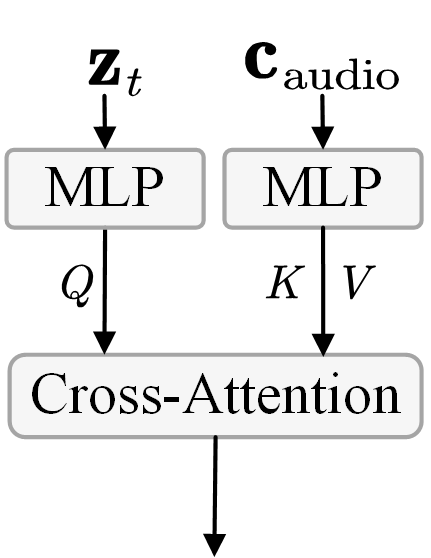}  
                    \caption{}  
                    \label{fig:AudioConditionCA}  
                \end{subfigure}%
                \vspace{-1mm}
                \caption{Different strategies of audio conditioning. (a) self-attention; (b) adaptive norm; (c) cross-attention.}
                \label{fig:AudioCondition}
        \end{minipage}
        \vspace{0.5cm} 
        \begin{minipage}[b]{\textwidth} 
            \centering
            \begin{subfigure}{0.22\linewidth} 
                \centering
                \includegraphics[width=0.95\linewidth]{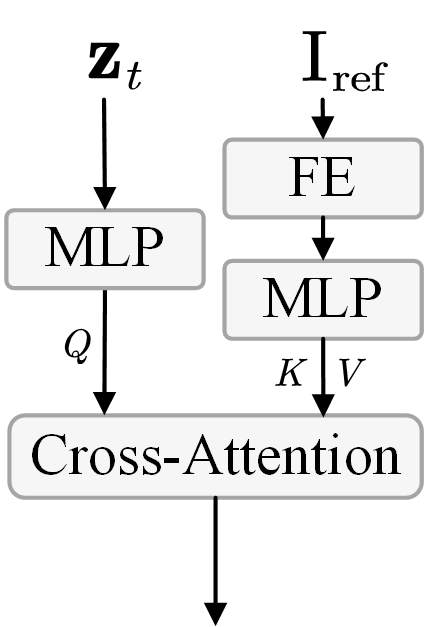} 
                \caption{}  
                \label{fig:RefConditionCA}  
            \end{subfigure}%
            \hfill
            \begin{subfigure}{0.22\linewidth} 
                \centering
                \includegraphics[width=0.95\linewidth]{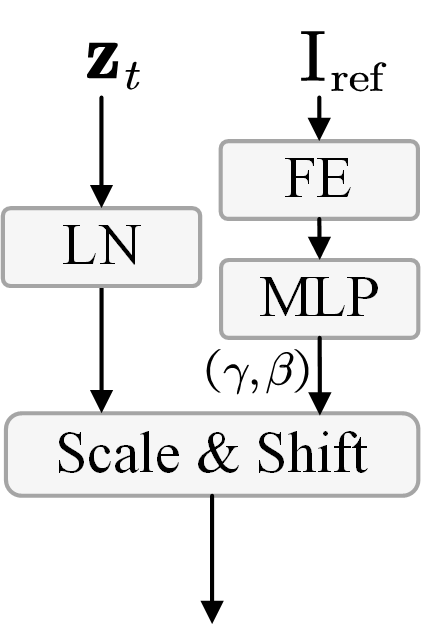}  
                \caption{}  
                \label{fig:RefConditionAda}  
            \end{subfigure}%
            \hfill
            \begin{subfigure}{0.23\linewidth} 
                \centering
                \includegraphics[width=0.95\linewidth]{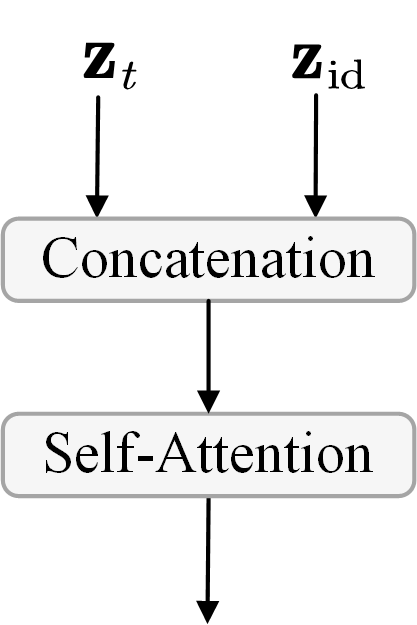}  
                \caption{}  
                \label{fig:RefConditionVAESA}  
            \end{subfigure}%
            \hfill
            \begin{subfigure}{0.30\linewidth} 
                \centering
                \includegraphics[width=\linewidth]{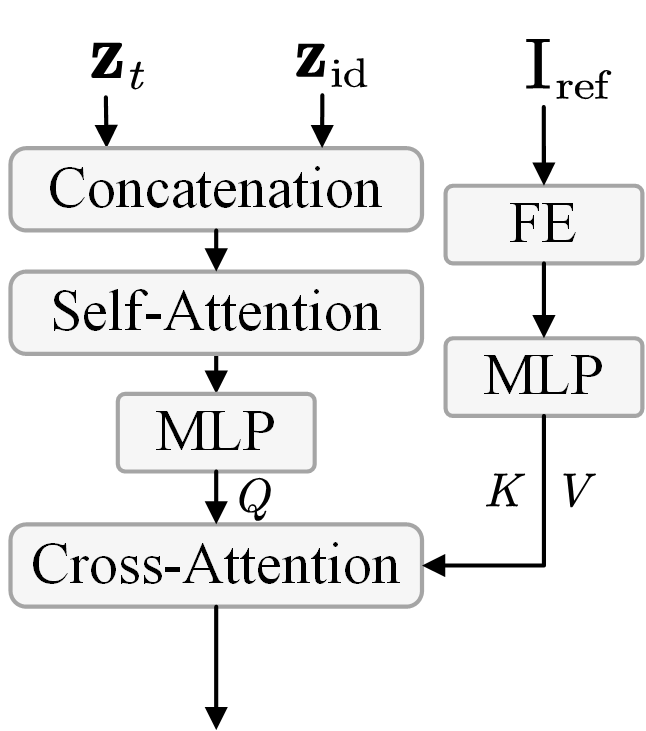}  
                \caption{}  
                \label{fig:RefConditionVAECA}  
            \end{subfigure}%
             \label{fig:RefCondition}
             \vspace{-1mm}
            \caption{{Different strategies of identity conditioning. FE refers to the face encoder. cross-attention demonstrates the best performance. (a) face attention; (b) face adaptive norm; (c) identity reference network; (d) face attention and identity reference network.}}
        \end{minipage}
    \end{minipage}
    \vspace{-5mm}
\end{figure*}

\section{Methodology}  
This methodology section systematically outlines the approaches employed in our study. 
Section~\ref{sec:baseline} describes the baseline transformer diffusion network, detailing its architecture and functionality. 
Section~\ref{sec:audio} focuses on the integration of speech audio conditions via a cross-attention mechanism. 
Section~\ref{sec:identity} discusses the implementation of the identity reference network, which is crucial for preserving facial identity coherence throughout extended video sequences. 
Section~\ref{sec:train} reviews the training and inference procedures used for the transformer diffusion network. 
Finally, Section~\ref{sec:data} details the comprehensive strategies for data sourcing and preprocessing.

\subsection{Baseline Transformer Diffusion Network}\label{sec:baseline}
\noindent\textbf{Baseline Network.}  
The CogVideoX model~\cite{yang2024cogvideox} serves as the foundational architecture for our transformer diffusion network, employing a 3D VAE for the compression of video data. 
In this framework, latent variables are concatenated and reshaped into a sequential format, denoted as \(\mathbf{z}_t\). 
Concurrently, the model utilizes the T5 architecture~\cite{raffel2023t5} to encode textual inputs into embeddings, represented as \(\mathbf{c}_{\text{text}}\). 
The combined sequences of video latent representations \(\mathbf{z}_t\) and textual embeddings \(\mathbf{c}_{\text{text}}\) are subsequently processed through an expert transformer network. 
To address discrepancies in feature space between text and video, we implement expert adaptive layer normalization techniques, which facilitate the effective utilization of temporal information and ensure robust alignment between visual and semantic data. 
Following this integration, a repair mechanism is applied to restore the original latent variable, after which the output is decoded through the 3D causal VAE decoder to reconstruct the video. 
Furthermore, the incorporation of 3D Rotational Positional Encoding (3D RoPE)~\cite{yang2024cogvideox} enhances the model's capacity to capture inter-frame relationships across the temporal dimension, thereby establishing long-range dependencies within the video framework.  

\noindent\textbf{Conditioning in Diffusion Transformer.}  
In addition to the textual prompt \(\mathbf{c}_{\text{text}}\), we introduce two supplementary conditions: the speech audio condition \(\mathbf{c}_{\text{audio}}\) and the identity appearance condition \(\mathbf{c}_{\text{id}}\).  

Within diffusion transformers, four primary conditioning mechanisms are identified: in-context conditioning, cross-attention, adaptive layer normalization (adaLN), and adaLN-zero~\cite{Peebles2022DiT}. 
Our investigation primarily focuses on cross-attention and adaptive layer normalization (adaLN). Cross-attention enhances the model's focus on conditional information by treating condition embeddings as keys and values, while latent representations serve as queries. 
Although adaLN is effective in simpler conditioning scenarios, it may not be optimal for more complex conditional embeddings that incorporate richer semantic details, such as sequential speech audio. Relevant comparative analyses will be elaborated upon in the experimental section.  

\subsection{Audio-Driven Transformer Diffusion}\label{sec:audio}
\noindent\textbf{Speech Audio Embedding.}  
To extract salient audio features for our proposed model, we utilize the wav2vec framework developed by Schneider et al.~\cite{schneider2019wav2vec}. The audio representation is defined as \(\mathbf{c}_{\text{audio}}\). 
Specifically, we concatenate the audio embeddings generated by the final twelve layers of the wav2vec network, resulting in a comprehensive semantic representation capable of capturing various audio hierarchies. 
This concatenation emphasizes the significance of phonetic elements, such as pronunciation and prosody, which are crucial as driving signals for character generation. 
To transform the audio embeddings obtained from the pretrained model into frame-specific representations, we apply three successive linear transformation layers, mathematically expressed as: $\mathbf{c}_{\text{audio}}^{(f)} = \mathcal{L}_3 \left( \mathcal{L}_2 \left( \mathcal{L}_1 \left( \mathbf{c}_{\text{audio}} \right) \right) \right)$, where \(\mathcal{L}_1\), \(\mathcal{L}_2\), and \(\mathcal{L}_3\) represent the respective linear transformation functions. This systematic approach ensures that the resulting frame-specific representations effectively encapsulate the nuanced audio features essential for the performance of our model.  

    

\noindent\textbf{Speech Audio Conditioning.}  
We explore three fusion strategies---self-attention, adaptive normalization, and cross-attention---as illustrated in Figure~\ref{fig:AudioCondition} to integrate audio condition into the DiT-based video generation model. Our experiments show that the cross-attention strategy delivers the best performance in our model. For more details, please refer to Section~\ref{sec:ablationstudy}. 

Following this, 
 we integrate audio attention layers after each face-attention layer within the denoising network, employing a cross-attention mechanism that facilitates interaction between the latent encodings and the audio embeddings. 
Specifically, within the DiT block, the motion patches function as keys and values in the cross-attention computation with the hidden states \(\mathbf{z}_t\): $\mathbf{z}_t = \text{CrossAttention}(\mathbf{z}_t, \mathbf{c}_{\text{audio}}^{(f)})$. This methodology leverages the conditional information from the audio embeddings to enhance the coherence and relevance of the generated outputs, ensuring that the model effectively captures the intricacies of the audio signals that drive character generation.

\begin{figure*}[!h]
    \centering
    
    \includegraphics[width=\linewidth]{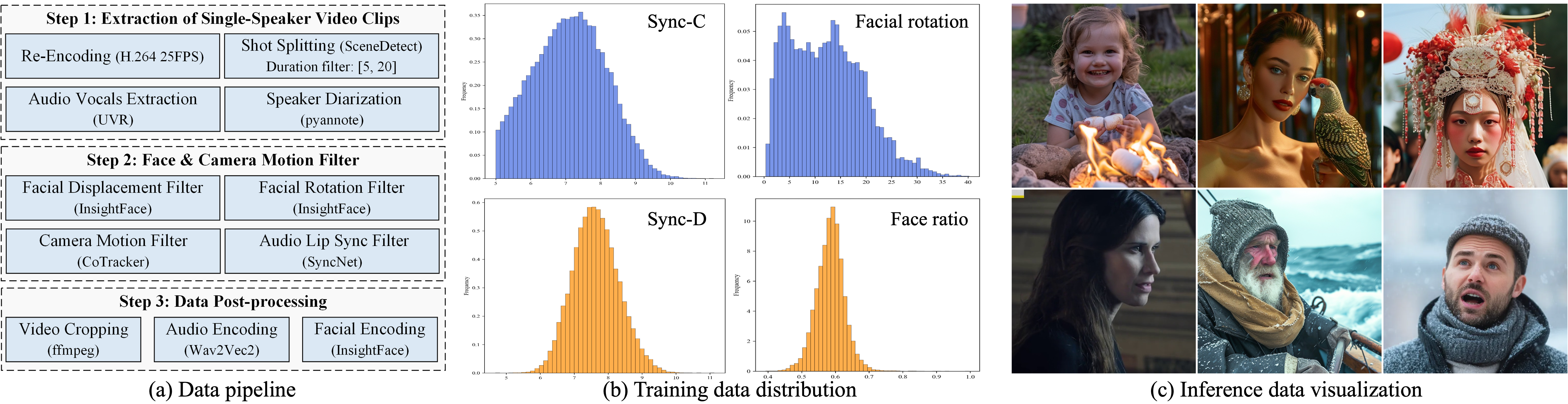}
    \vspace{-7mm}
    \caption{{
llustration of the dataset, including the flow of data processing, data distribution across different metric, and the visualization of inference data.}}
    \label{fig:data_statistics}
    \vspace{-6mm}
\end{figure*}

\subsection{Identity Consistent Transformer Diffusion}\label{sec:identity}



\noindent\textbf{Identity Reference Network.}  
Diffusion transformer-based video generation models encounter significant challenges in maintaining facial identity coherence, particularly as the length of the generated video increases. 
While incorporating speech audio embeddings as conditional features can establish a correspondence between audio speech and facial movements, prolonged generation often leads to rapid degradation of facial identity characteristics.  

To address this issue, we introduce a control condition within the existing diffusion transformer architecture to ensure long-term consistency of facial identity appearance. 
{We explore four strategies for appearance conditioning: 1) Face attention, where identity features are encoded by the face encoder and combined with a cross-attention module; 2) Face adaptive norm, which integrates features from the face encoder with an adaptive layer normalization technique; 3) Identity reference network, where identity features are captured by a 3D VAE and combined with some transformer layers; and 4) Face attention and Identity reference network, which encodes identity features using both the face encoder and 3D VAE, combining them with self-attention and cross-attention. Our experiments show that the combination with Face attention and Identity reference net achieves the best performance in our model. For further details, please refer to Section~\ref{sec:ablationstudy}. }

We treat a reference image as a single frame and input it into a causal 3D VAE to obtain latent features, which are then processed through a reference network consisting of 42 transformer layers. Mathematically, if \(\mathbf{I}_{\text{ref}}\) denotes the reference image, the encoder function of the 3D VAE is defined as: $\mathbf{z}_\text{id} = \mathcal{E}_{3D}(\mathbf{I}_{\text{ref}})$,
where \(\mathbf{z}_{\text{id}}\) represents the latent features associated with the reference image.  

During the operation of the reference network, we extract vision tokens from the input of the 3D full attention mechanism for each transformer layer, which serve as reference features \(\mathbf{z}_{\text{id}}\). These features are integrated into corresponding layers of the denoising network to enhance its capability, expressed as: $\mathbf{z}_{t, \text{enhanced}} = \text{SelfAttention}(\mathbf{z}_t, \mathbf{z}_{\text{id}}),$
where \(\mathbf{z}_t\) is the latent representation at time step \(t\). 
Given that both the reference network and denoising network leverage the same causal 3D VAE with identical weights and comprise the same number of transformer layers (42 layers in our implementation), the visual features generated from both networks maintain semantic and scale consistency. 
This consistency allows the reference network's features to incorporate the appearance characteristics of facial identity from the reference image while minimizing disruption to the original feature representations of the denoising network, thereby reinforcing the model's capacity to generate coherent and identity-consistent facial animations across longer video sequences.

\noindent\textbf{Temporal Motion Frames.}  
To facilitate long video inference, we introduce the last \(n\) frames of the previously generated video, referred to as motion frames, as additional conditions. Given a generated video length of \(L\) and the corresponding latent representation of \(l\) frames, we denote the motion frames as \(N\). 
The motion frames are processed through the 3D VAE to obtain \(n\) frames of latent codes. 
We apply zero padding to the subsequent \((l-n)\) frames and concatenate them with \(l\) frames of Gaussian noise. 
This concatenated representation is then patchified to yield vision tokens, which are subsequently input into the denoising network. By repeatedly utilizing motion frames, we achieve temporally consistent long video inference.

\subsection{Training and Inference}\label{sec:train}

\noindent\textbf{Training.}  
The training process consists of two phases: 
(1) In the first phase, we train the model to output videos with identity consistency. The parameters of the 3D VAE and face image encoder are fixed, while the parameters of the 3D full attention in the reference network, composed of a series of transformer blocks, and those in the denoising network, including the 3D full attention and face attention, are updated through training. The model's input comprises a randomly sampled reference image from the training video, a textual prompt, and the face embedding obtained from the face encoder, with the output being a sequence of 49 frames of video. 
(2) In the second phase, we train the model for audio-driven video generation. We insert audio attention modules into each transformer block of the denoising network, fixing the parameters of the other components and updating only those of the audio attention modules. The model's input consists of the reference image, input audio, and textual prompt, yielding a sequence of 49 frames of video.

\noindent\textbf{Inference.}  
During inference, the model receives a reference image, a segment of driving audio, a textual prompt, and motion frames as inputs. 
The model then generates a video that exhibits identity consistency and lip synchronization based on the driving audio. 
To produce long videos, we utilize the last two frames of the preceding video as motion frames, thereby achieving temporally consistent video generation.

\begin{table*}[th!]
    \centering
    \begin{minipage}{0.25\textwidth}
        \centering
        \resizebox{\textwidth}{!}{
        \begin{tabular}{c|c|c|c|c}
        \toprule
        \ &
          \multicolumn{1}{c|}{FID$\downarrow$} &
          \multicolumn{1}{c|}{FVD$\downarrow$} &
          \multicolumn{1}{c|}{Sync-C$\uparrow$} &
          \multicolumn{1}{c}{Sync-D$\downarrow$} \\ 
          \midrule
        SadTalker~\cite{zhang2022sadtalker}   & 22.340 & 203.860 & \textbf{7.885} & \textbf{7.545}  \\
        DreamTalk~\cite{ma2023dreamtalk}      & 78.147 & 790.660 & 6.376 & 8.364  \\
        AniPortrait~\cite{wei2024aniportrait} & 26.561 & 234.666 & 4.015 & 10.548  \\
        Hallo~\cite{xu2024hallo}              & 20.545 & 173.497 & 7.750 & 7.659  \\
        Ours                                  & \textbf{20.359} & \textbf{160.838} & 7.252 & 8.106  \\ 
        \midrule
        Real video                            & - & - & 8.700 & 6.597 \\ 
        \bottomrule
        \end{tabular}}
        \vspace{-2mm}
        \caption{Comparison with the other methods on HDTF dataset.} 
        \label{tab:comp_hdtf}
    \end{minipage}
    \hfill
    \begin{minipage}{0.30\textwidth}
        \centering
        \resizebox{\textwidth}{!}{
        \begin{tabular}{c|c|c|c|c|c}
        \toprule
            \ &
              \multicolumn{1}{c|}{FID$\downarrow$} &
              \multicolumn{1}{c|}{FVD$\downarrow$} &
              \multicolumn{1}{c|}{Sync-C$\uparrow$} &
              \multicolumn{1}{c|}{Sync-D$\downarrow$} &
              \multicolumn{1}{c}{E-FID$\downarrow$} \\ 
              \midrule
            SadTalker~\cite{zhang2022sadtalker}   & 50.015 & 471.163 & 6.922 & \textbf{7.921} & 95.194 \\
            DreamTalk~\cite{ma2023dreamtalk}      & 109.011 & 988.539 & 5.709 & 8.743 & 153.450 \\
            AniPortrait~\cite{wei2024aniportrait} & 46.915 & 477.179 & 2.853 & 11.709 & 88.986 \\
            Hallo~\cite{xu2024hallo}              & 44.578 & 377.117 & \textbf{7.191} & 7.984 & 78.495 \\
            Ours                                  & \textbf{43.271} & \textbf{355.272} & 6.527 & 9.113 & \textbf{71.210} \\ 
            \midrule
            Real video                            & - & - & 7.372 & 7.518 & - \\ 
            \bottomrule
        \end{tabular}}
        \vspace{-2mm}
        \caption{Comparison with other methods on Celeb-V dataset.}
        \label{tab:comp_celebv}
    \end{minipage}
    \hfill
    \begin{minipage}{0.39\textwidth}
        \centering
        \resizebox{\textwidth}{!}{
        \begin{tabular}{c|c|c|c|c|c|c}
        \toprule
        \ &
          \multicolumn{1}{c|}{Sync-C$\uparrow$} &
          \multicolumn{1}{c|}{Sync-D$\downarrow$} &
          \makecell{Subject\\Dynamic$\uparrow$} &
          \makecell{Background\\Dynamic$\uparrow$} &
          \makecell{Subject\\FVD$\downarrow$} &
          \makecell{Background\\FVD$\downarrow$} \\ 
          \midrule   
        SadTalker~\cite{zhang2022sadtalker}  & 3.845           & 10.378         & 2.953   & 0.220 & 470.377 &313.758 \\
        DreamTalk~\cite{ma2023dreamtalk}     & 4.498           & 11.005         & 6.958   & 1.806 & 835.480 &744.177 \\
        AniPortrait~\cite{wei2024aniportrait}&	1.685           & 12.025        & 3.351  & 1.769 & 473.173 & 302.716 \\
        Hallo~\cite{xu2024hallo}             & 4.654           & 10.202         & 5.268   & 1.272 & 394.627 & 291.052 \\
        Ours                                 & \textbf{6.154}  & \textbf{8.574} & 
         \textbf{13.286}  & \textbf{4.481} & \textbf{359.493} &\textbf{248.283} \\ 
        \bottomrule
        \end{tabular}}
        \vspace{-2mm}
        \caption{Comparison with other methods on our proposed wild dataset.}
        \label{tab:comparison_wild}
    \end{minipage}
    \vspace{-2mm}
\end{table*}

\begin{figure*}[t!]
    \centering
    \includegraphics[width=0.95\linewidth]{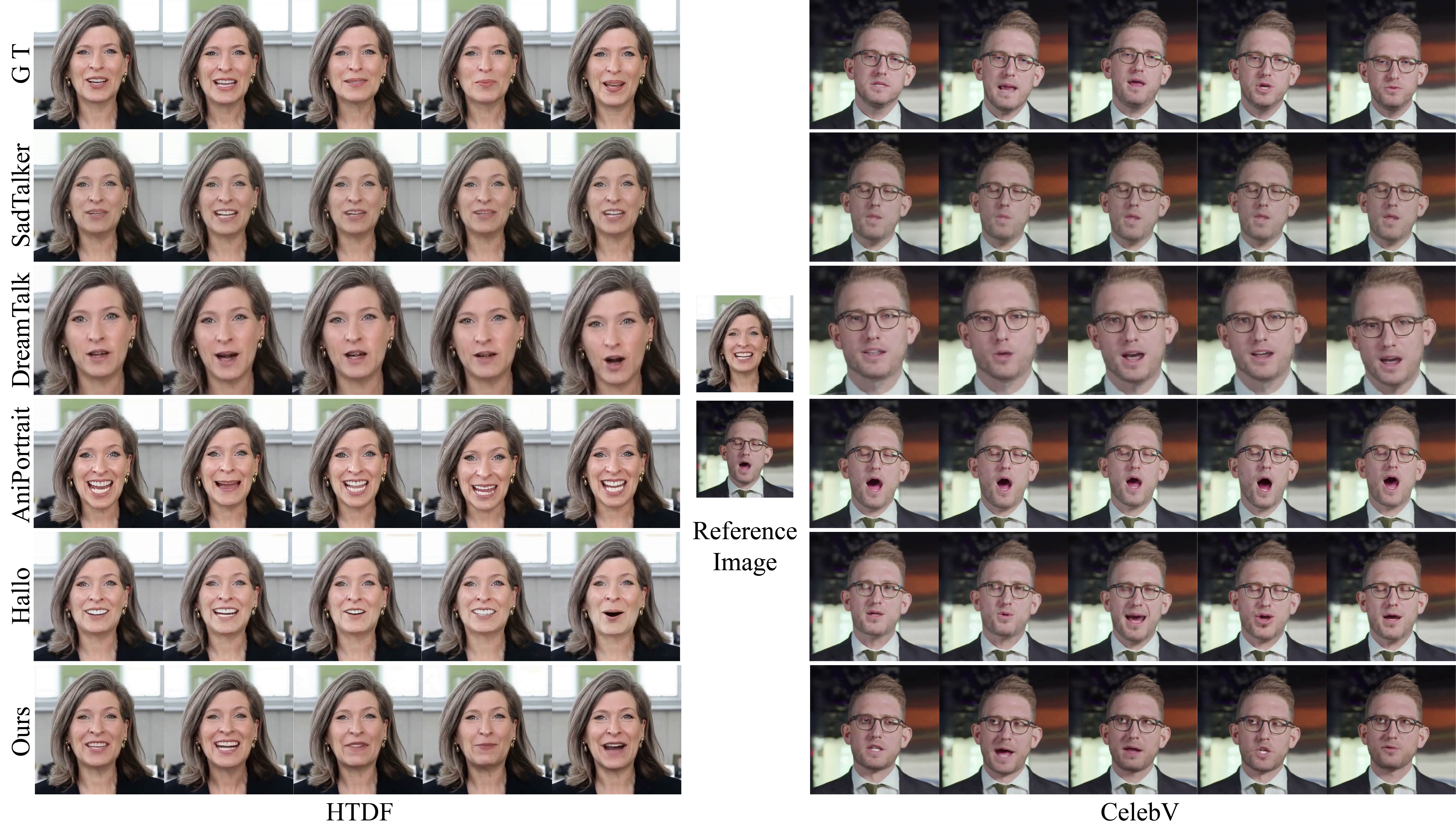}
    \vspace{-3mm}
    \caption{Qualitative comparison on the HTDF~(left) and CelebV~(right) data-set.}
    \vspace{-6mm}
    \label{fig:hdtf}
\end{figure*}

\subsection{Dataset}\label{sec:data}

In this section, we will give a detailed introduction of our data curation, including data sources, filtering strategy and data statistics.
Figure~\ref{fig:data_statistics} shows the data pipeline and the statistical analysis of the final data.

\noindent\textbf{Data Sources}
The training data used in this work is prepared from three distinct sources to ensure diversity and generalization. Specifically, the sources are: (1) HDTF dataset~\cite{zhang2021flow}, which contains 8 hours of raw video footage; (2) YouTube data, which consists of 1,200 hours of public raw videos; (3) a large scale movie dataset, which contains film videos of 2,346 hours. 
Our dataset contains a large scale of human identities and, however, we find that YouTube and movie dataset contains  a large amount of noised data. Therefore, we design a data curation pipeline as follows to construct a high-quality and diverse talking dataset, as shown in Figure~\ref{fig:data_statistics}a.


    
    

\noindent\textbf{Video Filtering.}
During the data pre-processing phase, we implement a series of meticulous filtering steps to ensure the quality and applicability of the dataset. The workflow includes three stages: extraction of single-speaker, motion filter and post-processing. Firstly, we select video of single-speaker. This stage aims to clean the video content to solve camera shot, background noise, etc, using existing tools~\cite{Plaquet23,Bredin23}. After that, we apply several filtering techniques to ensure the quality of head motion, head pose, camera motion, etc~\cite{karaev23cotracker,karaev24cotracker3,Chung16a}. In this stage, we compute all metric scores for each clip, therefore, we can flexibly adjust data screening strategies to satisfy different data requirement of our multiple training stages or strategies. Finally, based on the facial positions detected in previous steps, we crop the videos to a 3:2 aspect ratio to meet the model's input requirements. We then select a random frame from each video and use InsightFace~\cite{ren2023pbidr} to encode the face into embeddings, providing essential facial feature information for the model. Additionally, we extract the audio from the videos and encode it into embeddings using Wav2Vec2 model~\cite{baevski2020wav2vec}, facilitating the incorporation of audio conditions during model training.

\noindent\textbf{Data Statistics.}
Following the data cleaning and filtering processes, we conducted a detailed analysis of the final dataset to assess its quality and suitability for the intended modeling tasks. Finally, our training data contains about 134 hours training data, including 6 hours of high-quality data from HDTF dataset, 72 hours YouTube videos, and 56 hours movie videos. Figure~\ref{fig:data_statistics}b also shows other statistics, such as Lip Sync score~(Sync-C and Sync-D), face rotation, Face ratio~(the ratio of face height to video height).

\begin{figure*}[t!]
    \centering    \includegraphics[width=.90\linewidth]{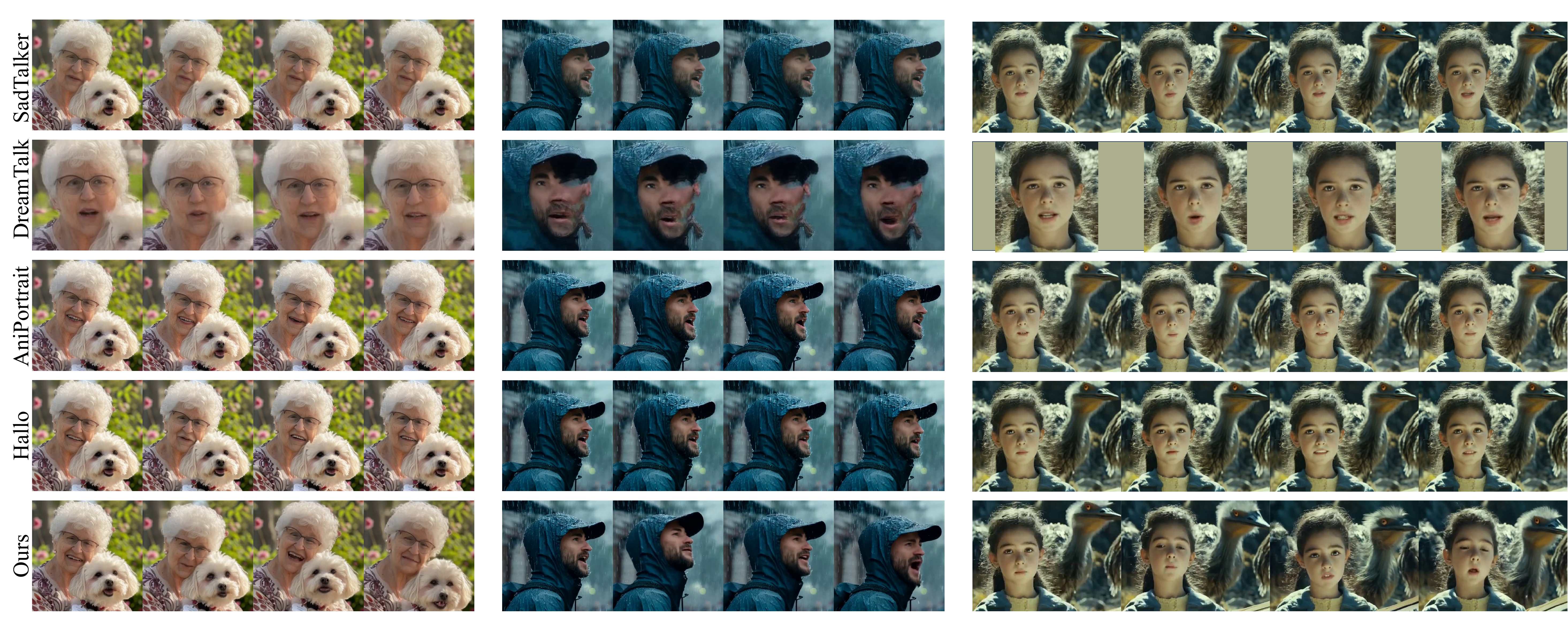}
    \vspace{-5mm}
    \caption{Complex facial identity with dynamic accessories subjects and different pose orientation.}
    
    \label{fig:portrait_complex_face}
    \vspace{-2mm}
\end{figure*}

\begin{figure*}[t!]
    \centering
    \includegraphics[width=0.9\linewidth]{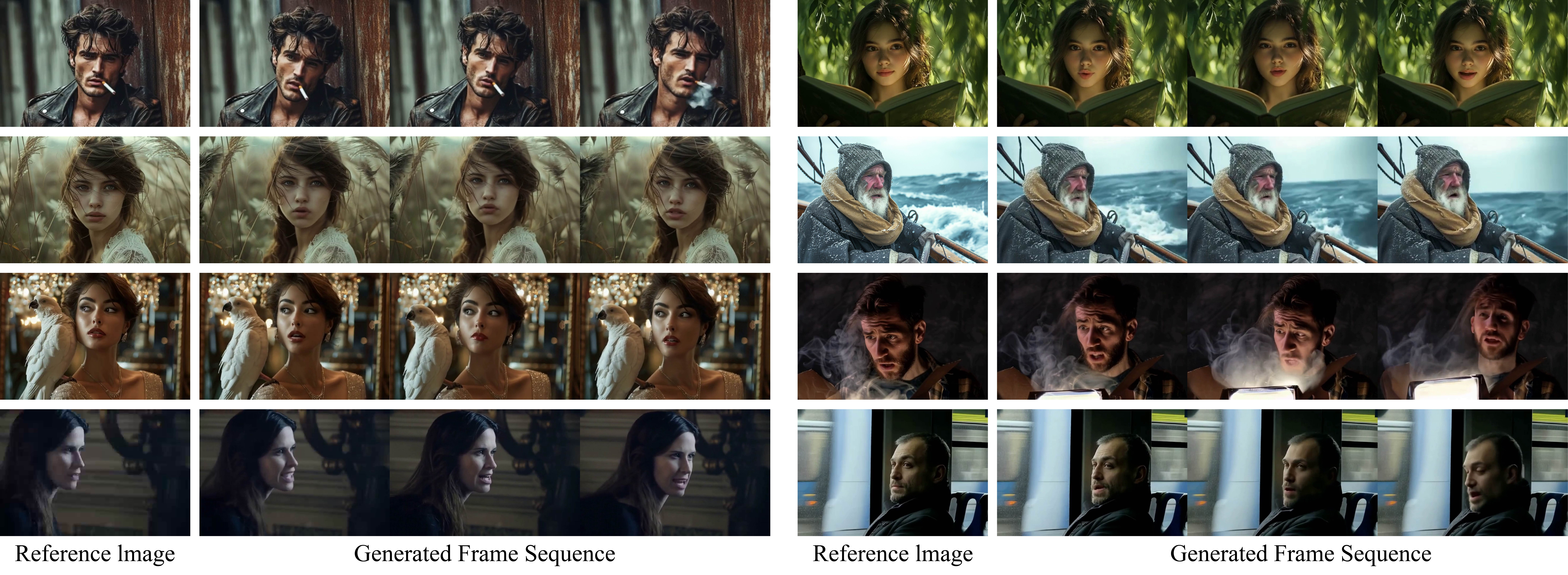}
    \vspace{-3mm}
    \caption{Complex scenes with dynamic foreground or background and various head poses.}
    \label{fig:ComplexScenes}
    \vspace{-5mm}
\end{figure*}



\section{Experiment}
\subsection{Experimental Setups}

\noindent\textbf{Implementation.}
{The model was trained using 64 NVIDIA A100 GPUs. In addition to training, all experiments were conducted on a GPU server equipped with 8 NVIDIA A100 GPUs. The first and second stages of the model were trained 20,000 steps respectively. During training, the batch size of each GPU was 1, and the learning rate was set to 1e-5. The resolution of the training video is 480 x 720, and it can generate a video with 49 frames at a time. In the training process, the audio embedding is dropped with a probability of 0.05, and the motion frames are randomly masked with a probability of 0.25.}



\subsection{Comparison with State-of-the-art}

\noindent\textbf{Comparison on HDTF and Celeb-V Dataset.}
{As shown in Table~\ref{tab:comp_hdtf} and ~\ref{tab:comp_celebv}, our method achieves best results on FID, FVD on both datasets. Although our approach shows some disparity compared to the state-of-the-art (SOTA) in lip synchronization, it still demonstrates promising results as illustrated in Figure~\ref{fig:hdtf}. This is because, to generate animated portraits from different perspectives, our training data primarily consists of talking videos with significant head and body movements, as well as diverse dynamic scenes, unlike static scenes with minimal motion. While this may lead to some performance degradation on lip synchronization, it better reflects realistic application scenarios. }


\begin{figure*}[!t]
    \centering
    \begin{minipage}{0.48\linewidth}
        \centering
        \includegraphics[width=\linewidth]{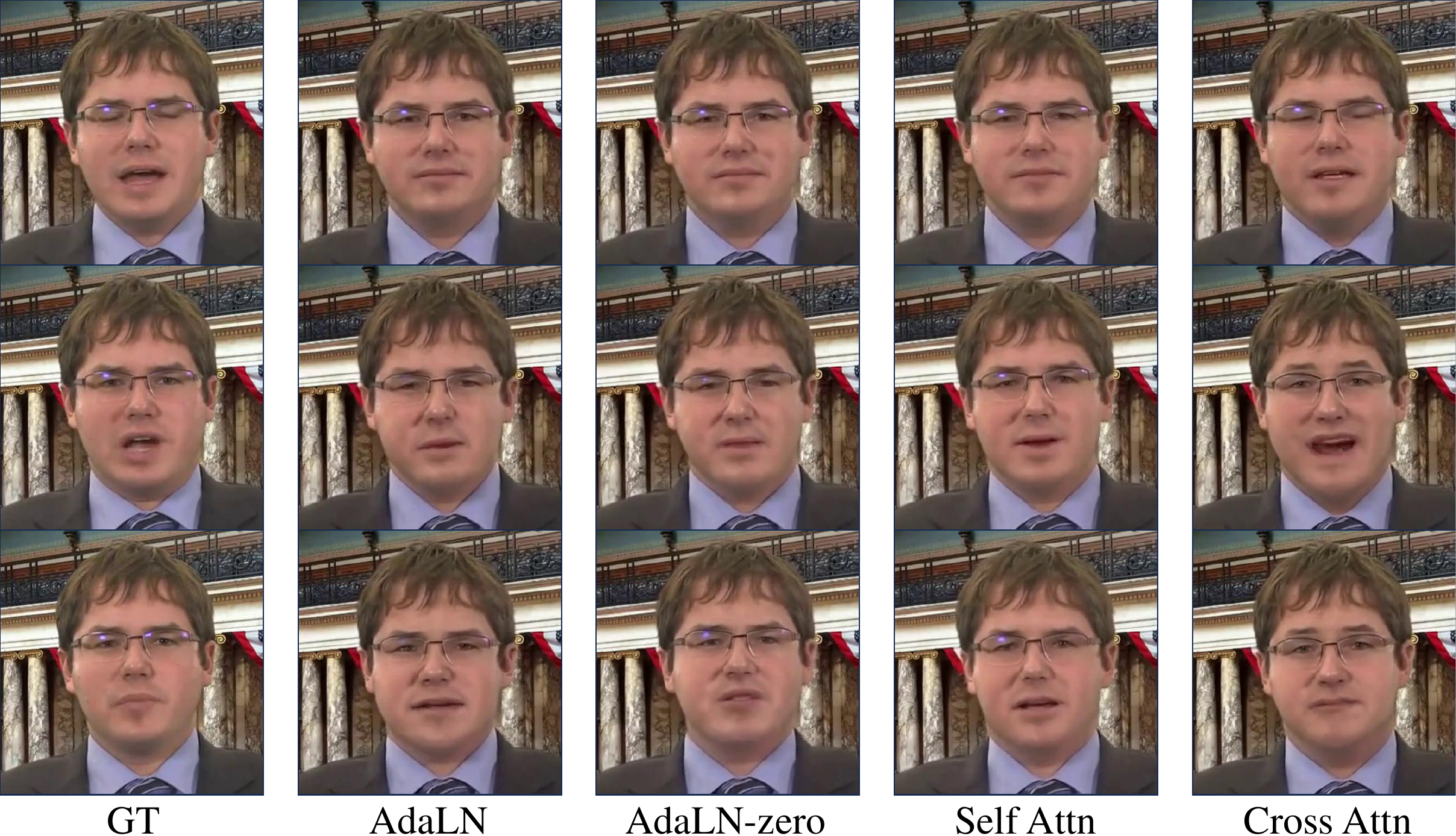}
        \vspace{-7mm}
        \caption{Qualitative comparison of different strategies for audio conditioning.}
        \label{fig:comparisn_of_audio_conditioning2}
    \end{minipage}
    \hfill
    \begin{minipage}{0.48\linewidth}
        \centering
        \includegraphics[width=.95\linewidth]{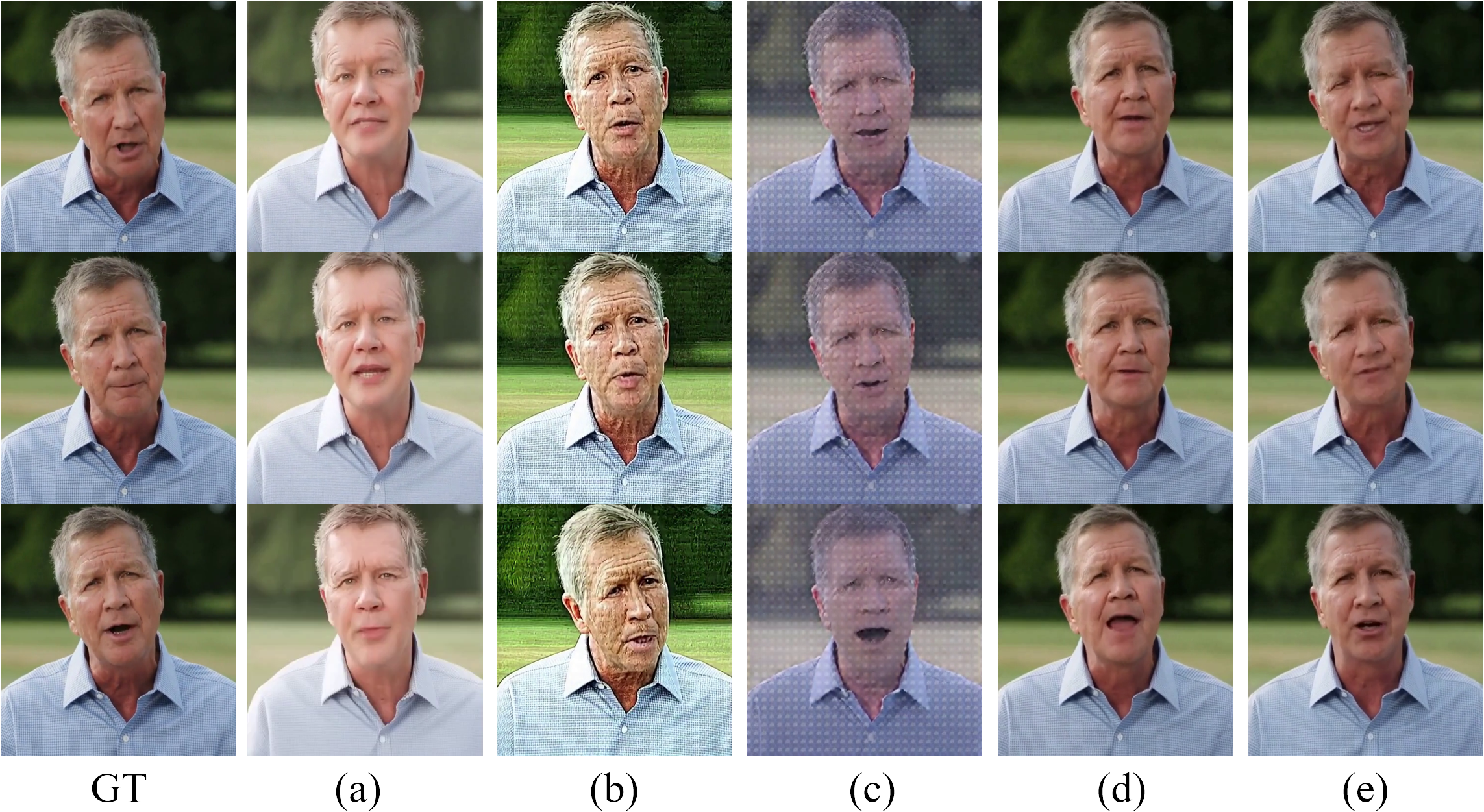}
        \caption{Qualitative comparison of different strategies for identity conditioning. The serial numbers correspond to those listed in Table~\ref{tab:identity_preserve}. }
        \label{fig:Comparison_of_the_appearance_reference_network}
    \end{minipage}
    \vspace{-4mm}
\end{figure*}

\noindent\textbf{Comparison on Wild Dataset.} 
To effectively demonstrate the performance of the general talking portrait video generation, we carefully collect 34 representative cases for evaluation. This dataset consists of portrait images with various head proportions, head poses, static and dynamic scenes and complex headwears and clothing. To achieve comprehensive assessment, we evaluate the performance on lip synchronization~(Sync-C and Sync-D), motion strength (subject and background dynamic degree) and  video quality~(subject and background FVD).
As shown in Table~\ref{tab:comparison_wild}, our method generates videos with largest head and background dynamic degree~(13.286 and 4.481) while keeping lip synchronization of highest accuracy. 

Figure~\ref{fig:portrait_complex_face} provides a qualitative comparison of different portrait methods on a ``wild" dataset. The results reveal that other methods struggle to animate side-face portrait images, often resulting in static poses or facial distortions. Additionally, these methods tend to focus solely on animating the face, overlooking interactions with other objects in the foreground---such as the dog next to the elderly, or the dynamic movement of the background---like the ostrich behind the girl. In contrast, as shown in Figure~\ref{fig:ComplexScenes} our method produces realistic portraits with diverse orientations and complex foreground and background scenes.

\begin{table}[t!]
\centering
\footnotesize
\begin{tabular}{c|c|c|c|c}
\toprule
  \makecell{Audio Injection Method} &
  \multicolumn{1}{c|}{FID$\downarrow$} &
  \multicolumn{1}{c|}{FVD$\downarrow$} &
  \multicolumn{1}{c|}{Sync-C$\uparrow$} &
  \multicolumn{1}{c}{Sync-D$\downarrow$} \\ 
  \midrule
 AdaLN       & 24.159 & 264.331 & 1.374 & 13.524 \\
 AdaLN-zero  & 24.029 & 276.403 & 1.398 & 13.553 \\ 
 Self Attn.  & 24.748 & 270.101 & 1.345 & 13.456 \\
 \midrule
 Cross Attn.(Ours) & \textbf{23.458} & \textbf{242.602} & \textbf{4.601} & \textbf{10.416} \\
 \bottomrule
\end{tabular}
\vspace{-2mm}
\caption{Comparison on the different strategy of audio conditioning.}
\label{tab:audio_injection}
\vspace{-3mm}
\end{table}

\begin{table}[t!]
\centering
\footnotesize
\resizebox{\linewidth}{!}{
\begin{tabular}{c|c|c|c|c|c}
\toprule
  \makecell{Identity Injection Method} &
  \multicolumn{1}{c|}{FID$\downarrow$} &
  \multicolumn{1}{c|}{FVD$\downarrow$} &
  \multicolumn{1}{c|}{Sync-C$\uparrow$} &
  \multicolumn{1}{c|}{Sync-D$\downarrow$} & 
  \multicolumn{1}{c}{ Subject consistency$\uparrow$} \\ 
  \midrule
 (a) No identity condition & 32.304 & 371.820 & 3.183 & 11.732 & 0.977 \\
 (b) Face attention & 57.541 & 740.536 & 4.042 & 10.682 & 0.974 \\
 (c) Face adaptive norm & 150.720 & 1587.395 & 3.822 & 12.324 & 0.904 \\
 (d) Identity reference network & 28.789 & 291.863 & 4.553 & \textbf{10.317} & 0.984 \\
 \midrule
 (e) Face attention and Identity reference network  & \textbf{23.458} & \textbf{242.602} & \textbf{4.601} & 10.416 & \textbf{0.988} \\ 
 \bottomrule
\end{tabular}
}
\vspace{-2mm}
\caption{Comparison of different identity injection method. ``No identity condition'' refers to the absence of any conditioning related to identity; ``Face attention'' and ``Face adaptive norm'' involve incorporating face embeddings using self-attention and adaptive layer normalization, respectively. ``Identity reference network'' refers to the introduction of identity features using a reference network.}
\label{tab:identity_preserve}
\vspace{-5mm}
\end{table}

\subsection{Ablation Study and Discussion}
\label{sec:ablationstudy}
\noindent\textbf{Audio Conditioning.} {Table~\ref{tab:audio_injection} and Figure~\ref{fig:comparisn_of_audio_conditioning2} illustrate the effects of various strategies for incorporating audio conditioning. The results demonstrate that using cross-attention to integrate audio improves lip synchronization by enhancing the local alignment between visual and audio features, particularly around the lips. This is evident from the improvements in Sync-C and Sync-D, and it also contributes to a degree of enhancement in video quality.}

\noindent\textbf{Identity Reference Network.} {
Table~\ref{tab:identity_preserve} and Figure~\ref{fig:Comparison_of_the_appearance_reference_network} evaluate different identity conditioning strategies.  The results indicate that without an identity condition, the model fails to preserve the portrait appearance. When using face embedding alone, the model introduces blur and distortion, as it focuses solely on facial features and disrupts the global visual context. To address this, we introduce an identity reference network to preserve global features while making facial motion more controllable through identity-based facial embeddings. Thus, the proposed method achieves a lower FID of 23.458 and FVD of 242.602, while maintaining lip synchronization.
}

\noindent\textbf{Temporal Motion Frames.} {Table~\ref{tab:motion_frame_num} presents an analysis of varying temporal motion frames. One motion frame achieves the highest Sync-C score (6.889) and the lowest Sync-D score (8.695), indicating substantial lip synchronization.}


\begin{table}[t!]
\centering
\footnotesize
\begin{tabular}{c|c|c|c|c}
\toprule
  \makecell{Motion Frame Number} &
  \multicolumn{1}{c|}{FID$\downarrow$} &
  \multicolumn{1}{c|}{FVD$\downarrow$} &
  \multicolumn{1}{c|}{Sync-C$\uparrow$} &
  \multicolumn{1}{c}{Sync-D$\downarrow$} \\ 
  \midrule 
 n = 1     & 24.040 & 242.708 & \textbf{6.889} & \textbf{8.695} \\
 n = 2     & \textbf{23.458} & \textbf{242.602} & 4.601 & 10.416 \\
 n = 4     & 24.459 & 269.904 & 5.109 & 10.489 \\
 n = 8     & 27.303 & 265.396 & 5.114 & 10.464 \\ 
 \bottomrule 
\end{tabular}
\vspace{-2mm}
\caption{Ablation on the number of motion frames.}
\label{tab:motion_frame_num}
\vspace{-6mm}
\end{table}

\section{Conclusion}
{
This paper introduces advancements in portrait image animation utilizing the enhanced capabilities of a DiT-based diffusion model. By integrating audio conditioning through cross-attention mechanisms, our approach effectively captures the intricate relationship between audio signals and facial expressions, achieving substantial lip synchronization. To preserve facial identity across video sequences, we incorporate an identity reference network. Additionally, we utilize motion frames to enable the model to generate long-duration video extrapolations. Our model produces animated portraits from diverse perspectives, seamlessly blending dynamic foreground and background elements while maintaining temporal consistency and high fidelity.
\paragraph{Acknowledgements.}
This project is sponsored by Natural Science Foundation of Shanghai under Grant No. 24ZR1407200 and Shanghai Oriental Talents Project under Grant No. QNKJ2024060.
}

{\small
\bibliographystyle{ieee_fullname}
\bibliography{egbib}
}

\end{document}


\title{Supplementary Materials: Highly Dynamic and Realistic Portrait Image Animation with Diffusion Transformer Networks}

\author{First Author\\
Institution1\\
Institution1 address\\
{\tt\small firstauthor@i1.org}
\and
Second Author\\
Institution2\\
First line of institution2 address\\
{\tt\small secondauthor@i2.org}
}
\maketitle

\section{Network}
\subsection{Identity Reference Network Details}

Maintaining subject consistency in diffusion transformer based video generation presents a significant challenge, particularly as video length increases. While integrating speech audio embeddings as conditional features aids in aligning facial movements with audio, prolonged generation often results in a degradation of facial identity fidelity.

To address this issue, we propose an identity reference network within the diffusion transformer framework designed to preserve facial identity coherence in realistic portrait animation. Figure~\ref{fig:differentref} illustrates various strategies for identity preservation after 10 seconds of video generation:

\noindent \textbf{(a) No Identity Condition.} In the absence of identity conditions, the model struggles to maintain adequate portrait coherence after 10 seconds.

\noindent \textbf{(b) Face Attention.} Incorporating features from the face encoder InsightFace~\cite{insightface2024} into the cross-attention module effectively captures high-level features—such as the appearance of age indicated by wrinkles in the reference image. However, this method still results in noticeable alterations to the subject's appearance.

\noindent \textbf{(c) Face Adaptive Norm.} Here, face embeddings obtained from InsightFace~\cite{insightface2024} are injected via an adaptive layer normalization technique. However, this approach also fails to preserve the subject's identity by emphasizing overall visual context at the expense of specific portrait features, potentially leading to distortion.

\noindent \textbf{(d) Identity Reference Network.} Our proposed identity reference network comprises several transformer blocks, each containing adaptive layer normalization layers, a 3D full attention layer, and a feed-forward layer. We first employ a 3D VAE to encode the reference image, then input these latent features into the identity reference network to extract reference image features. These features are concatenated with the input features of the 3D full-attention layer in the denoising network, allowing the reference image features to be injected through the 3D full-attention module. Our identity reference network effectively encodes reference images, preserving detailed identity and background features (e.g., the text “CPS”). However, it tends to introduce a smoothing effect that compromises finer details, such as wrinkles in the portrait.

\noindent \textbf{(e) Face Attention and Identity Reference Network.} Finally, we combine the identity reference network with the face encoder to incorporate higher-level semantic features. This integration enhances the portrait's characteristic attributes while maintaining identity fidelity.

\begin{figure*}[t!]
    \centering
    \includegraphics[width=0.98\linewidth]{figs/supp_2_4_image.pdf}
    \vspace{-3mm}
    \caption{Qualitative study of different setting of the identity injection method.
    Selected frames begin at 10s. (a) \textbf{No identity condition:} No specific conditions are applied to control the subject's appearance; (b) \textbf{Face attention:} Identity features from InsightFace~\cite{insightface2024} are processed through a cross-attention module; (c) \textbf{Face adaptive norm:} Identity features from InsightFace~\cite{insightface2024} are incorporated via adaptive layer normalization, applied through scaling and shifting.; (d) \textbf{Identity reference network:} Features are encoded using a reference network and integrated within the 3D full attention module.;(e) \textbf{Face attention and identity reference network:} Identity features are encoded separately via the InsightFace~\cite{insightface2024} and the reference network. These encoded features are then integrated within the 3D full attention module and processed through a cross-attention mechanism.}
    \label{fig:differentref}
    \vspace{-5mm}
\end{figure*}

\subsection{Training Details}

The training process comprises two phases: 

\noindent \textbf{(1) Identity Consistency Phase.} In this initial phase, we train the model to generate videos with consistent identity. The parameters of the 3D Variational Autoencoder (VAE) and face image encoder remain fixed, while the parameters of the 3D full attention blocks in both the reference and denoising networks, along with the face attention blocks in the denoising network, are updated during training. The model’s input includes a randomly sampled reference image from the training video, a textual prompt, and the face embedding. The textual prompt is generated using MiniCPM\cite{yao2024minicpm}, which describes human appearance, actions, and detailed environmental background. The face embedding is extracted via InsightFace\cite{insightface2024}. With these inputs, the model generates a video comprising 49 frames.

\noindent \textbf{(2) Audio-Driven Video Generation Phase.} In the second phase, we extend the training to include audio-driven video generation. We integrate audio attention modules into each transformer block of the denoising network, while fixing the parameters of other components and updating only those of the audio attention modules. Here, the model's input consists of a reference image, an audio embedding, and a textual prompt, resulting in a sequence of 49 video frames driven by audio.

\noindent \textbf{Implementation Details.}
We initialize the identity reference and denoising networks with weights derived from CogVideoX-5B-I2V\cite{yang2024cogvideox}. During both training phases, we employ the v-prediction diffusion loss\cite{salimans2022progressive} for optimization. Each training phase comprises 20,000 steps, utilizing 64 NVIDIA A100 GPUs. The batch size per GPU is set to 1, with a learning rate of \(1 \times 10^{-5}\). The resolution of the training videos is 480 x 720 pixels. To enhance video generation variability, the reference image, guidance audio, and textual prompt are dropped with a probability of 0.05 during training.

\begin{table*}[th!]
    \centering
    \resizebox{\linewidth}{!}{
    \begin{tabular}{c|ccc|c|c|c|c|c|c|c}
    \toprule
    &Audio & Text & Image & 
    Sync-C$\uparrow$ & Sync-D$\downarrow$ & 
    \makecell{Subject\\Dynamic$\uparrow$} & 
    \makecell{Background\\Dynamic$\uparrow$} & 
    \makecell{Subject\\FVD$\downarrow$} & 
    \makecell{Background\\FVD$\downarrow$} &
    \makecell{Subject\\Consistency$\uparrow$}\\ 
    \midrule
    $\lambda_t \downarrow$ &$\lambda_a = 3.5$ & $\lambda_t=1.0$ & $\lambda_i=1.0$ &  6.168 & 8.589 & 13.164 & 3.955 $\downarrow$ & 361.582 & 263.416 & 0.9813 \\ 
    Base &$\lambda_a=3.5$ & $\lambda_t=3.5$ & $\lambda_i=1.0$ &  6.154 & 8.574 & 13.286 & 4.481 & 359.493 & 248.283 & 0.9810 \\ 
    $\lambda_t \uparrow$ &$\lambda_a=3.5$ & $\lambda_t=6.0$ & $\lambda_i=1.0$ &  6.044 & 8.861 & 13.616 & 4.659 $\uparrow$ & 342.894 & 235.307 & 0.9808 \\ 
    $\lambda_a \uparrow$&$\lambda_a=6.0$ & $\lambda_t=3.5$ & $\lambda_i=1.0$ &  6.469 $\uparrow$ & 8.515 & 14.778 & 4.066 & 379.073 & 264.969 & 0.9809 \\  
    $\lambda_i \uparrow$&$\lambda_a=3.5$ & $\lambda_t=3.5$ & $\lambda_i=3.5$ & 6.023 & 8.654 & 12.599 & 4.219 & 367.225 & 265.414 & 0.9835 $\uparrow$ \\ 
    \bottomrule
    \end{tabular}}
    \vspace{-2mm}
    \caption{ Quantitative study of audio, text and image CFG scales on our proposed wild dataset. }
    \vspace{-4mm}
    \label{tab:abalation_cfg}
\end{table*}

\section{Experiments}

\subsection{Experimental Setup Details}
\paragraph{Evaluation Metrics.}
We employed a range of evaluation metrics for generated videos across benchmark datasets, including HDTF and Celeb-V. 
These metrics comprise Fréchet Inception Distance (FID)~\cite{Seitzer2020FID}, Fréchet Video Distance (FVD)~\cite{unterthiner2018towards}, Synchronization-C (Sync-C)~\cite{Chung16a}, Synchronization-D (Sync-D)~\cite{Chung16a}, and E-FID~\cite{tian2024emo}. 
FID and FVD quantify the similarity between generated images and real data, while Sync-C and Sync-D assess lip synchronization accuracy. E-FID evaluates image quality based on features extracted from the Inception network.

Besides, we introduced V-bench~\cite{huang2023vbench} metrics to enhance evaluation, focusing on dynamic degree and subject consistency. 
Dynamic degree is measured using RAFT~\cite{teed2020raft} to quantify the extent of motion in generated videos, providing a comprehensive assessment of temporal quality.
Subject consistency is measured through DINO feature similarity, ensuring uniformity of a subject's appearance across frames.


\paragraph{Baseline Approaches.}
We considered several representative audio-driven talking face generation methods for comparison, all of which have publicly available source code or implementations. These methods include SadTalker~\cite{zhang2022sadtalker}, DreamTalk~\cite{ma2023dreamtalk}, AniPortrait~\cite{wei2024aniportrait}, and Hallo~\cite{xu2024hallo,cui2024hallo2}. 
The selected approaches encompass both GANs and diffusion models, as well as techniques utilizing intermediate facial representations alongside end-to-end frameworks. 
This diversity in methodologies allows for a comprehensive evaluation of the effectiveness of our proposed approach in comparison to existing solutions.

\subsection{Ablation and Discussion} 

\paragraph{CFG Scales for Diffusion Model.}  
Table~\ref{tab:abalation_cfg} provides a quantitative analysis of video generations using various CFG scales for audio, text, and reference images. A comparison between the second and fourth rows demonstrates that increasing the audio CFG scale enhances the model's ability to synchronize lip movements. The text CFG scale significantly influences the video’s dynamism, as indicated in the first three rows, where both the subject's and the background's dynamics increase with higher text CFG scales. Conversely, the reference image CFG scale primarily governs the subject's appearance; higher values improve subject consistency, as illustrated by the second and fifth rows. Among the tested configurations, setting \(\lambda_a=3.5\), \(\lambda_t=3.5\), and \(\lambda_i=1.0\) yields a balanced performance. This interplay between visual fidelity and dynamics underscores the effectiveness of CFG configurations in generating realistic portrait animations.

\begin{figure*}[th!]
    \centering
    \includegraphics[width=1.0\linewidth]{figs/interact_human.pdf}
    \caption{Condition on Interactive Subjects. The white number represents the BLIP score, which measures the alignment between the generated video and the textual prompts. A higher value indicates a better alignment. Our method achieves alignment comparable to that of CogVideX, maintaining the controllability of interactive subjects even after introducing the audio condition.}
    \label{fig:inter}
\end{figure*}

\begin{figure*}[th!]
    \centering
    \includegraphics[width=1.0\linewidth]{figs/2-5-2.pdf}
    \caption{Textual Condition on Foreground and Background. The white number represents the BLIP score, which measures the alignment between the generated video and the textual prompts. A higher value indicates a better alignment. Our method achieves alignment comparable to that of CogVideX, maintaining the controllability of foreground and background after incorporating the audio condition.}
    \label{fig:fgbg}
\end{figure*}

\subsection{Generation Controllability} 

\noindent\textbf{Textual Prompt for Subject Animation.}
To evaluate whether textual conditional controllability is effectively preserved, we conducted a series of experiments comparing the performance of our method to that of the baseline model, CogVideoX~\cite{yang2024cogvideox}, using same text prompts. As shown in Figure~\ref{fig:inter}, the white number represents the BLIP score, which measures how well the generated videos align with the textual prompts. A higher score indicates better alignment. The results shows that our model maintains its ability for textual control, achieving a BLIP score comparable to that of CogVideX, and effectively captures the interaction between different subjects as dictated by the textual prompts.

\noindent\textbf{Textual Prompt for Foreground and Background Animation.} We also explore model's ability to follow the foreground and background textual prompt. As illustrated in Figure~\ref{fig:fgbg}, our method animates the foreground and background subjects naturally, such as the ocean waves and flickering candlelight. The results demonstrates the model's ability to control foreground, and background with the textual caption, which is maintained even after introducing the audio condition.




\subsection{Limitations and Future Works}
Despite the advancements in portrait image animation techniques presented in this study, several limitations warrant acknowledgment. 
While the proposed methods improve identity preservation and lip synchronization, the model's ability to realistically represent intricate facial expressions in dynamic environments still requires refinement, especially under varying illumination conditions. 
Future work will focus on enhancing the model's robustness to diverse perspectives and interactions, incorporating more comprehensive datasets that include varied backgrounds and facial accessories. 
Furthermore, investigating the integration of real-time feedback mechanisms could significantly enhance the interactivity and realism of portrait animations, paving the way for broader applications in live media and augmented reality.

\section{Safety Considerations}
The advancement of portrait image animation technologies, particularly those driven by audio inputs, presents several social risks, most notably concerning the ethical implications associated with the creation of highly realistic portraits that may be misused for deepfake purposes. 
To address these concerns, it is essential to develop comprehensive ethical guidelines and responsible use practices.
Moreover, issues surrounding privacy and consent are prominent when utilizing individuals' images and voices. It is imperative to establish transparent data usage policies, ensuring that individuals provide informed consent and that their privacy rights are fully protected.
By acknowledging these risks and implementing appropriate mitigation strategies, this research aims to promote the responsible and ethical development of portrait image animation technology.

{\small
\bibliographystyle{ieee_fullname}
\bibliography{egbib}
}